# An enhanced neural network based approach towards object extraction

S.K. Katiyar                                                                 P.V. Arun

Department Of Civil Engineering
MA National Institute of Technology, India

ABSTRACT

The improvements in spectral and spatial resolution of the satellite images have facilitated the automatic extraction and identification of the features from satellite images and aerial photographs. An automatic object extraction method is presented for extracting and identifying the various objects from satellite images and the accuracy of the system is verified with regard to IRS satellite images. The system is based on neural network and simulates the process of visual interpretation from remote sensing images and hence increases the efficiency of image analysis. This approach obtains the basic characteristics of the various features and the performance is enhanced by the automatic learning approach, intelligent interpretation, and intelligent interpolation. The major advantage of the method is its simplicity and that the system identifies the features not only based on pixel value but also based on the shape, haralick features etc of the objects. Further the system allows flexibility for identifying the features within the same category based on size and shape. The successful application of the system verified its effectiveness and the accuracy of the system were assessed by ground truth verification.

## 1. INTRODUCTION

The availability of high resolution images has facilitated the automatic extraction and identification of objects and is not only demanding but also a task of practical importance. Object based approach has been applied to various contexts namely map updation, classification, automatic registration, change detection, target recognition, urban planning, site modelling etc. Efficiencies of various methodologies in this context are affected by different factors such as scene complexity, feature variability, undefined feature geometry, and sensor resolution (Bruno, 1998; Lari, 2011).

There are a number of object extraction algorithms and most of them are specific to features to be extracted and adopts computationally complex methods for the extraction (Shuichi, 2011). Neural Network (NN) technique that models the human brain has the ability to derive meaning from complicated or imprecise data. This ability of NN is exploited to extract patterns and detect trends that are too complex to be noticed by either humans or other computer techniques. Bruno et al. (Bruno, 1998) suggested a road extraction system in which tests segments (matched filters for short road segments) was adopted to detect the road position. The method was not completely automatic since starting point and direction was to be provided. Barzohar et al **(1996)** discussed a road extraction system which linkers, correlation trackers, and region based followers for detection. Road tracing based on profile matching and kalman filtering was introduced by Vosselman et al (1995) how ever seemed to be noise sensitive especially in case of small distances between the profiles. Model based road extraction from images proposed by Carsten (1998) also failed due to the lack of topology consideration as well as context. A semiautomatic feature extraction based on snakes was suggested by John et al (1986) however the method needs manual involvement and the snakes seems to get distorted in the proximity of other features in the image. Gruen (1997) suggested a road extraction method based on wavelet transformed image and least square matching but seemed to be less efficient for the non linear features. A map-based semantic modelling for the extraction of objects was proposed by Quint et al (1995) but the efficiency was limited by the mere use of line segments for image analysis and accuracy was dependent on the availability of the map at hand.

Wang et al (2009) proposed a fuzzy mask based method in order to perform road pixels extraction in which a road membership value is attributed to each pixel. Lau Bee Theng et al (2008), in his methods adopted active contour model to extract objects like land parcels and buildings from high resolution satellite imageries. A technique that integrates the advantages of both fuzzy theory and Hopfield type neural network for object extraction from noisy background was proposed by Bhattacharyya et al (1998). Many approaches based on neural networks for road detection did not achieve significant improvements as they relied on a small context (9×9 pixels being the largest) for prediction and used very little training data (Boggess, 1993). An automatic road extraction system using a neural network with millions of trainable weights was suggested by Mnih et al (2010) how ever considered only pixel values as parameter for the extraction.

Most of the errors in the current systems are due to the ambiguous nature of the labelling task. Porway et al. (2008) used a grammar to model relationships between objects such as cars, trees, and roofs for the purpose of parsing aerial images. An object extraction method was proposed Jiangye Yuan et al(2009) based on locally excitatory globally inhibitory oscillator networks (LEGION) in which oscillation network can segment the image and extract objects by the scene and object feature. A knowledge-based method for automatic road extraction from high resolution remotely sensed images was proposed by Trinder et al (2003). They described road structures as a generalized anti parallel pair and their radiometric and geometric properties are expressed as rules in Prolog by virtue of which the extraction was accomplished. But this method was based on the static learning approach, ie system was unable to update the set of properties for road as a new one is encountered.

In this paper we propose a neural network based method for feature extraction which can extract the various features in to separate layers. The adoption of neural network for the feature extraction has many added advantages as error tolerance, non-linear programming, facility for training etc. as compared to the other approaches .The method can be effectively used for the various GIS systems as that used for planning and other purposes. The system automates the image analysis and simulates the visual interpretation of images along with enhancing the process with speed, simplicity, accuracy, intelligent interpretation and automatic interpolation. The system as implemented in mat lab is computationally simple and faster than its back propagation based counter parts implemented in c and other languages. The method is distinguishable from the existing methods which are based on the neural net due to the fact that it takes in to account the shape, haralick features of the objects in addition to its pixel value and also it uses back propagation which is more reliable than other training methods. The method also provides intelligent interpolation and intelligent interpretation which enhance its capability to detect objects accurately.

Our objective is to suggest a neural network based method for the automatic extraction of features from remotely sensed images after taking in to consideration various factors as shape, pixel intensity, context etc of the feature for its extraction.  The system should also provide intelligent interpolation in the sense that it should be able to remove the breaks, gaps etc. in the features also it should provide intelligent interpretation when the features are ambiguous. In this paper we propose a system based on feed forward neural net trained by back propagation algorithm. The haralick texture factors are also considered for detection and system has a rule set stored in prolog DB.

## 2. METHOD

The Satellite images namely PAN, CARTOSAT, and Google earth images of Bhopal area are used as test images from which the various features are extracted. The images are first subjected to contrast adjustment by adjusting the histogram and the edges are detected by canny edge detection operator to detect the boundaries of various common features and in case of bright features erosion followed by dilation is applied .The processed image is then fed to the trained neural net to extract or to select the features and the selected features are again subjected to curve fitting. The features outside the required boundary are eliminated by setting the pixel as black. The Haralick's texture features are also sued along with the pixel value and shape to identify the features .The haralick features include Energy, Correlation, Inertia, Entropy, Inverse Difference Moment, Sum Average, Sum Variance, Sum Entropy, Difference Average, Difference Variance, Difference Entropy, Information measure of correlation 1 and correlation 2.The extraction of specific features is done by using SOM networks and prolog rule set. The system uses a set of rules stored in a prolog data base for interpreting ambiguous features and identifying them. The process is summarised schematically in fig.1.

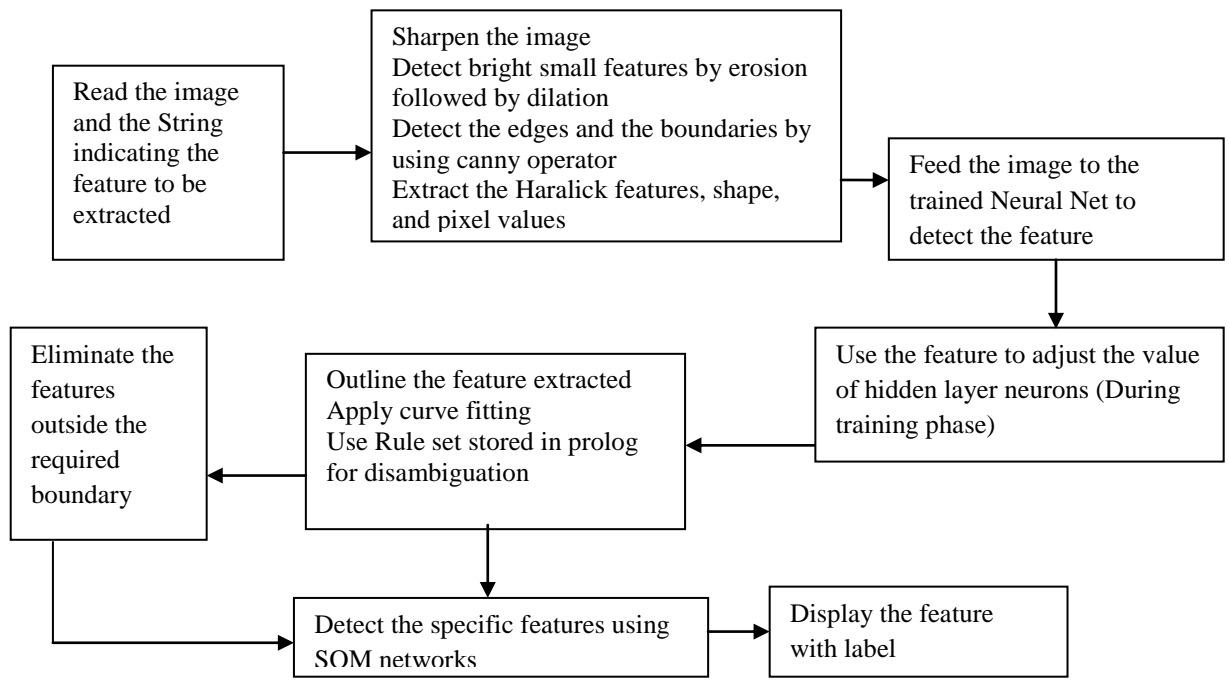

Fig.1. Schematic representation of NN method

### 2.1 Training of Neural net for feature detection and extraction

The neural net specific for each feature is first trained by giving the various example feature images each as column vectors of an input matrix. Each column of the input matrix is a feature vector which we want to find to which class this feature belongs. For each of these feature vectors their distance from the desired feature vectors is calculated and the weights of the hidden layers are adjusted or the network is said to be learned. The SOM network is also trained to make it enable for detection.

### 2.2 Feature Detection using trained neural net

The image from which feature to be extracted is flattened, i.e. converted to a column vector form and fed to the feature specific neural net which finds the group similar to the desired feature and the similarity defined by the error value for each feature class. The column vector form corresponding to each image is the resultant of pixel values as well as the haralick features corresponding to that image. The shape of the feature will be considered by using a prolog DB and modifying the column vector value accordingly. The boundary is set over the feature detected and the rest of the pixels are made black and the pixels within the boundary to white. The pixel values, shape, haralick features of the objects are considered for the identification of specific features using SOM networks [38].

## 3. Results

The investigations of this research work revealed that considerable success have been achieved with the procedure. The adopted NN approach effectively detected various features better than the existing methodologies and accuracy is revealed from the extraction of roads, rivers, buildings etc. over the study areas. The efficiency of the traditional classifying approaches with reference to the adopted NN approach has been evaluated using the various statistical measures and the results are as summarised in table 2. The ground truthing is done with reference to the Google earth and Differential Global Positioning System (DGPS) survey over the study area using Trimble R3 DGPS equipment. The increased value of Kappa statistics and over all accuracy indicates better method.

| S.No | Sensor | Methodology | Kappa statistics | Overall Accuracy (%) |
|---|---|---|---|---|
| 1 | LISS 3 | Mahalanobis | 0.93 | 93.13 |
| 2 | LISS 3 | Minimum Distance | 0.92 | 94.58 |
| 3 | LISS 3 | Maximum Likelihood | 0.96 | 96.83 |
| 4 | LISS 3 | Parrellelepipid | 0.95 | 96.81 |
| 5 | LISS 3 | Feature Space | 0.97 | 95.15 |
| 6 | LISS 3 | NN Approach | 0.98 | 96.12 |
| 8 | LISS 4 | Mahalanobis | 0.90 | 91.40 |
| 9 | LISS 4 | Minimum Distance | 0.91 | 93.00 |
| 10 | LISS 4 | Maximum Likelihood | 0.94 | 94.80 |
| 11 | LISS 4 | Parrellelepipid | 0.93 | 94.62 |
| 12 | LISS 4 | Feature Space | 0.94 | 95.31 |
| 13 | LISS 4 | NN | 0.96 | 96.81 |

The performances of these methodologies have been evaluated by comparing the areal extents of various features. The features having well defined geometry like lakes, parks etc are selected for the comparative analysis. The original surface areas of the features are calculated by manual digitization using ERDAS and comparative the results are presented in the Table 3. Comparative analyses of the areal extents also indicate that the CNN approach yields better results compared to the other methods.

| S.No | Sensor | Feature | Reference Area(km²) | Methodology | Areal Extent(km²) |
|---|---|---|---|---|---|
| 1 | LISS3 | Lake | 32.5 | Mahalanobis | 25.42 |
| | | | | Minimum Distance | 24.31 |
| | | | | Maximum Likelihood | 27.37 |
| | | | | Parallelepiped | 28.58 |
| | | | | Feature Space | 26.82 |
| | | | | NN Approach | 28.01 |
| 2 | LISS3 | Parks | 2.13 | Mahalanobis | 0.82 |
| | | | | Minimum Distance | 0.89 |
| | | | | Maximum Likelihood | 1.45 |
| | | | | Parallelepiped | 1.37 |
| | | | | Feature Space | 1.56 |
| | | | | NN Approach | 2.07 |
| 3 | LISS4 | Lake | 32.81 | Mahalanobis | 24.31 |
| | | | | Minimum Distance | 23.40 |
| | | | | Maximum Likelihood | 25.12 |
| | | | | Parallelepiped | 26.24 |
| | | | | Feature Space | 27.17 |
| | | | | NN | 28.01 |
| 4 | LISS4 | Parks | 2.37 | Mahalanobis | 0.51 |
| | | | | Minimum Distance | 0.72 |
| | | | | Maximum Likelihood | 1.53 |
| | | | | Parallelepiped | 1.14 |

|  |  |  |  | Feature Space | 1.46 |
|  |  |  |  | NN | 1.62 |

The system showed improved performance in extracting various features of Bhopal area from PAN, CARTOSAT, Google earth images. The outputs of the system over various images with specific to various features are discussed below. The investigation regarding the accuracy of the feature extraction process over various satellite images revealed that considerable success was achieved with the procedure.

### 3.1 Extraction of road network

The system was tested for extracting the road networks from the PAN, Google Earth, LANDSAT image sensor image for new market area of Bhopal city and the results are as shown if Fig 2-4. The intermediate image of extraction from PAN image is shown in Fig 2(a) and the extracted network is shown in Fig 2(c). The whole road network of the area was successfully extracted and breaks due to shadows and obstacles where successfully eliminated by intelligent interpolation as evident from the results as shown below.

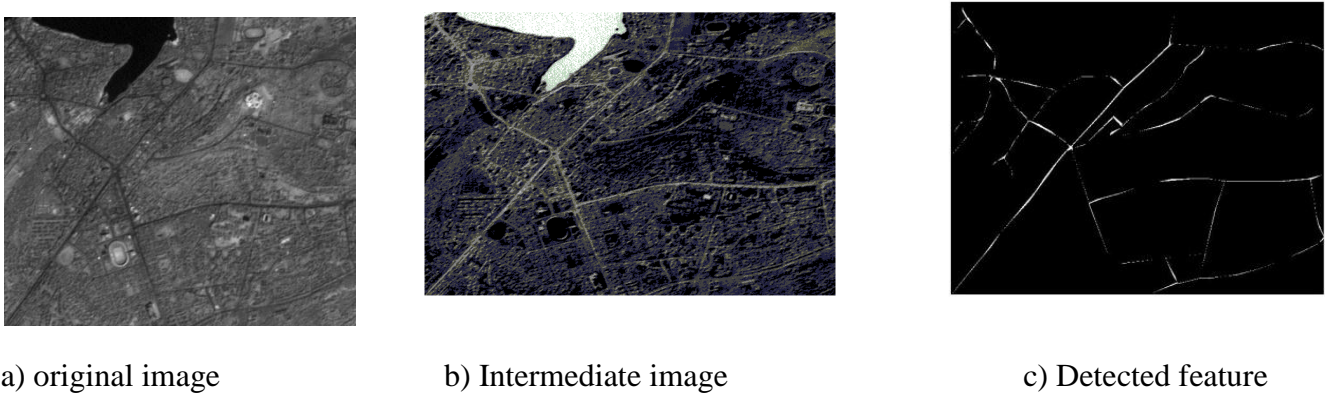

a) original image　　　　　　　　b) Intermediate image　　　　　　　　c) Detected feature

Fig 2: Road network extracted from PAN sensor image

The road network extracted from Google earth image is as shown below

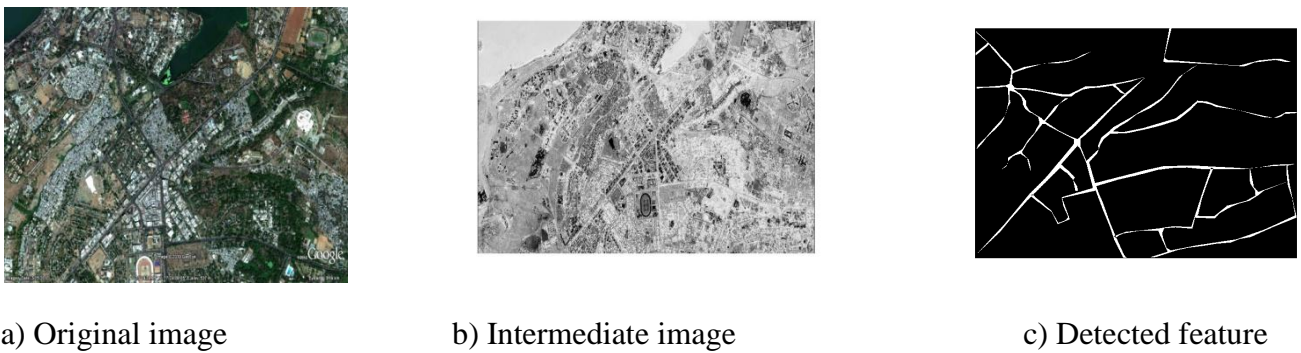

a) Original image　　　　　　　　b) Intermediate image　　　　　　　　c) Detected feature

Fig 3: Road network extracted from Google Earth image

The system provides option for extracting high ways,local roads etc seperately based on width and other parameters and the result of road network extracted based on width from Google earth image is as shown

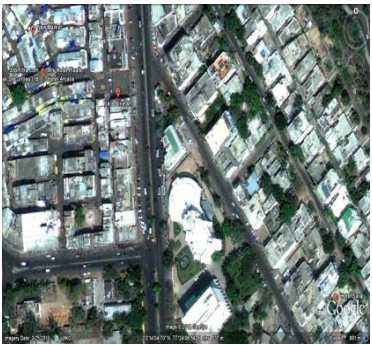 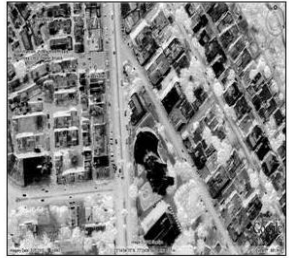 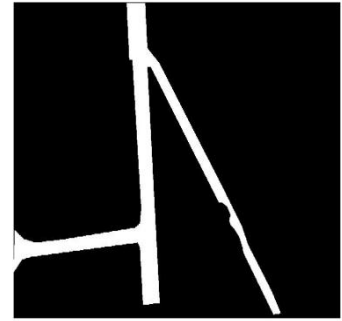

a) Original image    b) Intermediate image    c) Detected feature

Fig 4: Road network extracted from Google Earth image

The aacuracy of the features extracted were verified by ground truth verification and field visits

### 3.2 Extraction of lake from PAN image

The lake extracted from the PAN sensor image of Bhopal is also shown below,the steps where as discussed in the algorithm,the intermediate result is as shown in Fig 5(a) and the extracted water body is shown in Fig 5(c).

### 3.3 Extraction of water bodies from LANDSAT image

The water bodies extracted from the LANDSAT TM image of Wardha river basin is also shown in fig 6., context sensitivity of the system and use of haralick features help to extract rivers alone .The outputs can be used for detecting the changes in river flow over a period of years.The extracted water bodies are as shown below

### 3.4 Extraction of vehicle from Google Earth image

The one of the advanced feature of the system in extracting specific features is evidently shown in extracting each vehicle seperately from a google earth image as shown in fig.7.Here system has made use of the rules stored in prolog database for differentiating small buildings from vehicle.Also a specific vehicle was selected based on the size and shape parameter.

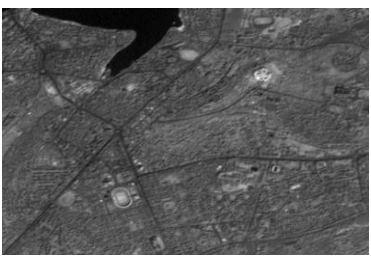 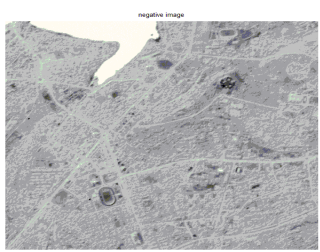 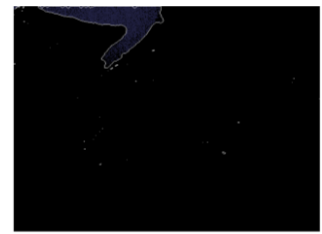

a) Original image    b) Intermediate image    c) Detected feature

Fig 5: Water body extracted from PAN sensor image

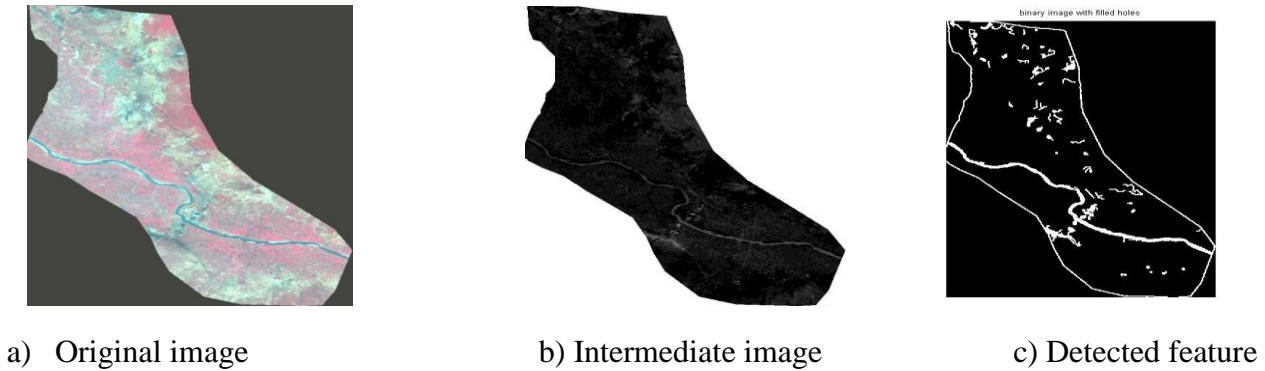

a)  Original image                     b) Intermediate image                     c) Detected feature

Fig 6: Water body extracted from Landsat sensor image

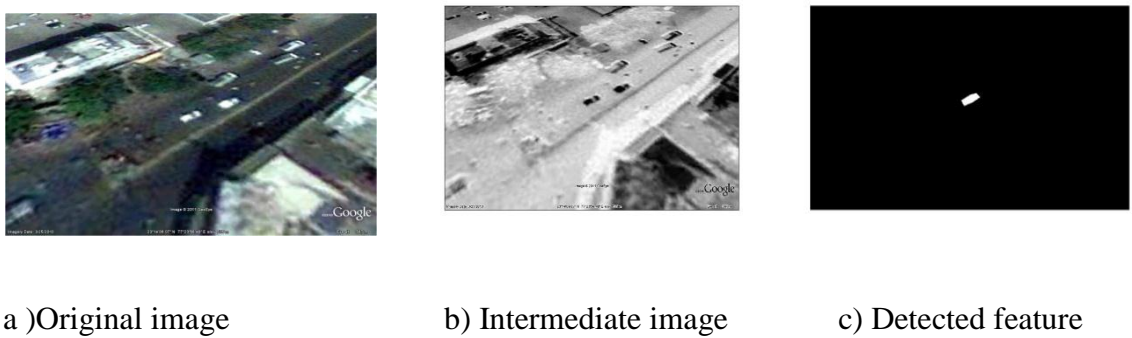

a )Original image                      b) Intermediate image                     c) Detected feature

Fig 7: Vehicle extracted from Google Earth image

## 4. CONCLUSIONS

The intelligent interpretation of the system using rule set stored in prolog and intelligent interpolation using the curve fitting makes the system superior to the existing methods. The system identifies the features correctly as it uses the haralick features, shape of the objects in addition to pixel values for the object extraction. The system was able to accurately extract the various features from the given satellite images (LISS III, LISS IV, PAN etc)and the accuracy of the system was tested by ground truth verification. The results have shown that the system out performs the existing methods in accuracy and the number of features that can be extracted. The system as implemented in Mat lab software is computationally simple and faster than its back propagation based counter parts implemented in c and other languages. The future improvements of the system include enhancement of the system by the use of context information and general rules for image analysis.